\documentclass[letterpaper]{article} 
\usepackage[]{aaai2027}
\usepackage[hyphens]{url}  
\usepackage{graphicx} 
\usepackage{natbib}  
\usepackage{caption} 
\nocopyright

\usepackage{algorithm}
\usepackage{algorithmic}

\usepackage{amsmath}
\usepackage{amssymb}
\usepackage{multirow}
\usepackage{array}

\usepackage{newfloat}
\usepackage{listings}
\DeclareCaptionStyle{ruled}{labelfont=normalfont,labelsep=colon,strut=off} 
\floatstyle{ruled}
\newfloat{listing}{tb}{lst}{}
\floatname{listing}{Listing}

\usepackage{booktabs}

\title{Calibration of Large Model Quantization based on HiFloat4 Format}
\author{
Jing Huang,
Yu Cheng,
Ziwei Yu,
Lan Zhao,
Sitan Li,
Guangxiu Yao,
Xin Wang\corresponding,
Yuanyong Luo,
Hengyi Liang,
Ting Li,
Bo Liu,
Hu Liu,
Michael Bi Mi,
Heng Liao
}
\affiliations{
    Huawei \\
    \textsuperscript{*}Email: wangxin237@huawei.com

}

\begin{document}

\maketitle

\begin{abstract}
Quantization of Large Language Models (LLMs) has becoming increasingly important for reducing the computational costs and resource requirements for practical deployment. Growing interest in this area has motivated the community to move beyond 8-bit quantization towards 4-bit precision, leading to the development of customized 4-bit formats such as MXFP4 and NVFP4. Recently, HiFloat4 (HiF4) has emerged as a competitive alternative which adopts a Block Floating-Point (BFP) format with a three-level scaling hierarchy, thereby achieving a significantly larger dynamic range. While quantization can substantially reduce memory usage and accelerate inference, it frequently introduces accuracy degradation. Calibration is therefore essential for narrowing the performance gap between low-precision quantized models and their full-precision counterparts. Nevertheless, existing calibration methods are mainly designed and validated using integer-precision formats, such as INT4, lacking support for HiF4 due to its hierarchical BFP structure. In this paper, we investigate three calibration approaches, namely HiGPTQ, HiAutoRound, and HiFlatQuant, to further enhance model quantization based on the HiF4 format. Extensive experiments demonstrate that all three proposed approaches improve inference accuracy compared with direct HiF4 conversion, reducing the gap between HiF4 and BF16 to less than 1\% and outperforming the state-of-the-art NVFP4 format across various models and downstream tasks.
\end{abstract}


\section{Introduction}

Recent years have witnessed a surge in Large Language Models (LLMs) with increasing model sizes and widespread deployment across diverse applications \cite{naveed2025review_LLMs}. Consequently, the trade-offs among model performance, memory usage, and deployment cost have received substantial attention, particularly for resource-constrained deployment scenarios \cite{zhou2024survey_efficient_inference}. As a result, model quantization has emerged as a critical approach for reducing memory consumption and improving computational throughput by converting high-precision model parameters into low-bit numerical formats \cite{lang2024quantization_review}. Research on model quantization is mainly advanced through two complementary directions: designing low-bit data formats to improve numerical expressiveness \cite{rouhani2023microscaling} and developing calibration algorithms to reduce accuracy degradation caused by low-precision representation \cite{yang2025survey_NN_quantization}.

Recently, 8-bit formats such as FP8 \cite{nvidia_h100_FP8}, MXFP8 \cite{amd_mi350x_MXFP8, rouhani2023microscaling} and HiF8 \cite{xu_huawei_HiF8, luo2024ascend_HiF8} have been deployed in commercial products, motivating further investigation into even higher compression rates. As a result, 4-bit formats have emerged as a promising direction for further reducing model sizes. For example, MXFP4 \cite{rouhani2023microscaling} uses a block size of 32, representing each element with a 4-bit E2M1 format and assigning an E8M0 scaling factor to each block. Here, in the E$X$M$Y$ notation, $X$ denotes the number of exponent bits, while $Y$ represents the number of mantissa bits. Similarly, NVFP4 \cite{abecassis2025pretraining_NVFP4} adopts the same E2M1 format for each element, but uses an E4M3 floating-point scale and reduces the block size to 16, enabling a more fine-grained representation. More recently, HiFloat4 (HiF4) \cite{luo2026hifloat4} has emerged as a competitive alternative, which introduces a three-level hierarchical scaling mechanism with a block size of 64, capturing both inter- and intra-group dynamic ranges within the limited 4-bit representational space. As a result, HiF4 achieves a broader dynamic range and higher accuracy while also reducing hardware cost.

While low-bit data formats define the underlying representation schemes, model quantization refers to the process of converting high-precision model parameters, such as FP32, FP16 or BF16, into low-precision formats \cite{yang2025survey_NN_quantization}. Model quantization can be classified into two categories: Quantization-Aware Training (QAT) and Post-Training Quantization (PTQ) \cite{zhu2024compression_survey}. QAT refers to training models with quantization in mind, enabling the models to adapt to the effect of quantization during training. In contrast, PTQ directly quantizes a full-precision model without the requirement for retraining. While QAT frequently achieves better results, its dependence on model retraining requires substantial training data and computational resources, making PTQ a more practical choice for large-scale deployment \cite{zhou2024survey_efficient_inference}.

Despite its benefits in reducing memory usage and accelerating inference, quantization inevitably introduces quantization error, which reflects the discrepancy between the quantized model and the high-precision counterpart \cite{sun2024FlatQuant}. To mitigate this error, various calibration techniques have been proposed within the PTQ framework, including GPTQ \cite{frantar2022GPTQ}, AutoRound \cite{cheng2024AutoRound} and FlatQuant \cite{sun2024FlatQuant}. These methods improve the quantization process through different calibration mechanisms, reducing the accuracy gap and enhancing downstream task performance. Nevertheless, these approaches are primarily designed and validated for the naive integer-based formats, such as INT4, and cannot be directly applied to HiF4 due to its hierarchical BFP structure. 

To address this research gap, this work proposes three HiF4-based quantization calibration approaches, namely HiGPTQ, HiAutoRound, and HiFlatQuant. Experimental results demonstrate that, compared with vanilla HiF4 conversion, the proposed approaches further improve quantization accuracy, achieving better performance across various downstream tasks and surpassing the state-of-the-art NVFP4-based quantization. The main contributions of this work are summarized as follows:

\begin{itemize}
    \item We propose HiGPTQ, HiAutoRound, and HiFlatQuant, three calibration algorithms for HiF4-based LLM quantization.
    \item We conduct extensive experiments across various LLMs and downstream tasks, demonstrating the effectiveness of the proposed approaches in enhancing HiF4-based quantization.
\end{itemize}

\section{Background}

This section introduces the HiF4 data format and reviews three existing calibration techniques, which together provide the foundation for the proposed methodologies.

\subsection{HiFloat4}

\begin{figure}[t]
    \centering
    \includegraphics[width=\linewidth]{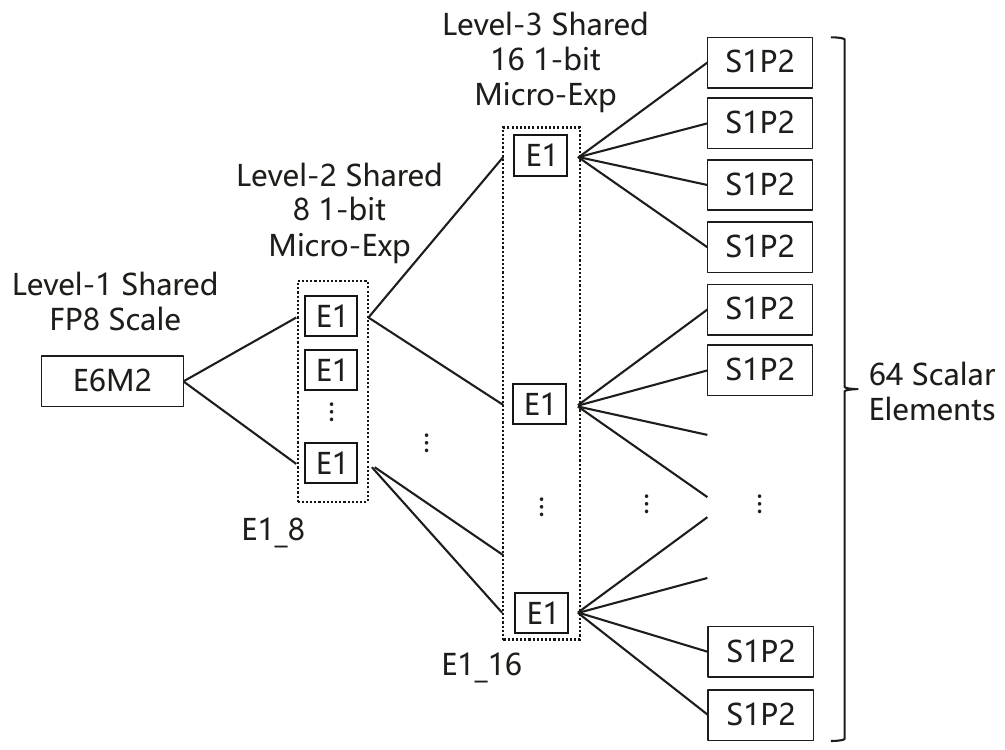}
    \caption{HiFloat4 Block Floating-Point (BFP) data structure.}
    \label{fig:hif4}
\end{figure}

HiF4 \cite{luo2026hifloat4} represents an emerging 4-bit Block Floating-Point (BFP) format that employs a three-level scaling hierarchy with a block size of 64. Figure \ref{fig:hif4} illustrates the structure of a HiF4 block. The first-level scale is represented by an FP8 value, with 6 bits allocated to the exponent and 2 bits to the mantissa, denoted as E6M2. HiF4 uses an exponent bias of 48 and an implicit integer bit of 1 in the mantissa field. As a result, the first-level scale has a maximum value of $1.50 \times 2^{15}$ and a minimum value of $1.00 \times 2^{-48}$. Additionally, the E6M2 component encodes NaN (Not a Number) when all 8 bits are set to 1.

The second- and third-level scales adopt 8-way and 16-way 1-bit micro-exponents, denoted as E1\_8 and E1\_16. Each micro-exponent directly encodes either 1 or 0 for a fine-grained power-of-two scaling. Under this hierarchy, each level-3 group contains 4 elements, while each level-2 group contains two level-3 groups. Overall, the three levels of scaling metadata require 32 bits, corresponding to an overhead of 0.5 bits per element.

The in-group elements are represented using a 4-bit sign-magnitude S1P2 format. In this notation, $S$ denotes the sign bit, the following 1 bit represents the integer part, and $P2$ represents two-bit fractional part. Consequently, the S1P2 component can represent a maximum value of $1.75$ and a minimum value of $0.25$, with the corresponding signed values ranging from $\pm0.25$ to $\pm1.75$. When the 1-bit integer part and 2-bit fractional parts are all zero, the value is represented as zero. 

Let $\{S1P2\}_i$ denote the $i$-th element within a block size of 64, $\{E1\_8\}_j$ denote the $j$-th micro-exponent in the 8-way level-2 scaling, $\{E1\_16\}_k$ denote the $k$-th micro-exponent in the 16-way level-3 scaling. Each value $V_i$ where $i \in [1, 64]$, in a HiF4 block can be represented as follows. If $\text{E6M2}=\text{NaN}$, then $V_i=\text{NaN}$ for all $i \in [1, 64]$. Otherwise, $V_i$ can be calculated as follows:

\begin{equation}
V_i =
\mathrm{E6M2}
\times
2^{\left(
\{E1\_8\}_{\lceil i/8 \rceil}
+
\{E1\_16\}_{\lceil i/4 \rceil}
\right)}
\times
\{S1P2\}_i
\end{equation}

Readers are referred to the HiF4 white paper \cite{luo2026hifloat4} for a more comprehensive explanation.


\subsection{Large Model Quantization}

This section reviews three representative quantization techniques, including GPTQ, AutoRound, and FlatQuant, which optimize the quantization process through different mechanisms.

\subsubsection{GPTQ}

GPTQ performs weight quantization in a layer-wise manner and formulates the quantization of each layer as a reconstruction problem based on the principle of error compensation.
Specifically, given a set of calibration data, GPTQ first collects the input activations $X$ for each layer and computes the Hessian matrix as $\mathbf{H}=2\mathbf{X}\mathbf{X}^T + \lambda \mathbf{I}$, where $\lambda$ is a small value for numerical stability. It then computes the inverse Hessian $\mathbf{H}^{-1}$ and applies Cholesky decomposition to obtain factors used for efficient error compensation during quantization. GPTQ subsequently splits the weight matrix into blocks of size $B$ and quantizes the weights block by block. For each block, it initializes an error matrix $\mathbf{E} \leftarrow \mathbf{0}_{d_{row}\times d_{col}}$, and performs quantization along the columns of the block. After quantizing each column, GPTQ computes the corresponding quantization error and updates the remaining unquantized weights in the same block to compensate for this error, using the pre-computed Cholesky factors. 
This process is described in Eqs.~\eqref{eq:GPTQ_error}--\eqref{eq:GPTQ_block_weight_update}, where $i=\{0, B, 2B, ...\}$ denotes the starting column index of each block and $j=\{i, ..., i+B-1\}$ denotes the column index within the current block.

\begin{equation}
\mathbf{E}_{:,j-i} \leftarrow \frac{\mathbf{W}_{:,j} - \operatorname{quant}(\mathbf{W}_{:,j})}{\left[ \mathbf{H}^{-1} \right]_{jj}}
\label{eq:GPTQ_error}
\end{equation}

\begin{equation}
\mathbf{W}_{:,\,j:(i+B)}
\leftarrow
\mathbf{W}_{:,\,j:(i+B)} - \mathbf{E}_{:,\,j-i} \cdot H_{j, j:(i+B)}^{-1}
\label{eq:GPTQ_block_weight_update}
\end{equation}

GPTQ adopts a lazy batch-update strategy: only after a block has been fully processed, it updates $\mathbf{H}^{-1}$ and $\mathbf{W}$ globally, thereby improving the efficiency of the quantization process. Specifically, the update of $\mathbf{H}^{-1}$ is realized by removing the corresponding rows from the Cholesky factors matrix, while the update of $\mathbf{W}$ is realized according to Eq.~\eqref{eq:GPTQ_global_weight_update}.

\begin{equation}
\mathbf{W}_{:,(i+B):}
\leftarrow
\mathbf{W}_{:,(i+B):} - \mathbf{E} \cdot \mathbf{H}_{i:(i+B), (i+B):}^{-1}
\label{eq:GPTQ_global_weight_update}
\end{equation}

As GPTQ quantizes weights column by column, it is not directly applicable to HiF4: processing each column independently could not fully exploit the hierarchical scaling structure of HiF4 and may lead to suboptimal quantization results. To address this limitation, we propose HiGPTQ, a calibration method tailored to the BFP-based data formats. 


\subsubsection{AutoRound}

AutoRound \cite{cheng2024AutoRound} also formulates weight quantization as a layer-wise reconstruction problem. 
Its main idea is to introduce learnable parameters to control how high-precision weights are rounded to their corresponding low-precision values, thereby reducing the reconstruction error. For integer-based quantization, the quantization scale is computed from the weight range. AutoRound introduces two learnable parameters, $\alpha$ and $\beta$, to automatically adjust the upper and lower bounds used for scale computation, as shown in Eq.~\eqref{eq:autoround_scale}

\begin{equation}
\label{eq:autoround_scale}
s = \frac{max(\mathbf{W}) \ast \alpha - min(\mathbf{W}) \ast \beta}{2^{bit} - 1}
\end{equation}

In addition, AutoRound learns a rounding matrix $\mathbf{V}$ to adjust the rounding direction of each weight element. The quantized weight matrix, $\widetilde{\mathbf{W}}$, can be obtained as follows:

\begin{equation}
\widetilde{\mathbf{W}} = s \ast \operatorname{clip} \left( \left\lfloor \frac{\mathbf{W}}{s} + zp + \mathbf{V} \right\rceil,\,
n,\, m \right), \quad n,m \in \mathbb{N}
\end{equation}

\noindent where $zp$ denotes the zero point, and $n$ and $m$ denote the lower and upper bounds of the quantized integer range, respectively. 

The learnable parameters $\alpha$, $\beta$, and $\mathbf{V}$ are optimized together by minimizing the layer reconstruction error:

\begin{equation}
\min_{\alpha, \beta, \mathbf{V}} \left\| \mathbf{W}\mathbf{X} - \widetilde{\mathbf{W}}\mathbf{X} \right\|_F^2
\end{equation}

AutoRound adopts SignSGD optimizer, which updates parameters solely based on the sign of the gradient. This design reduces sensitivity to large variations in gradient magnitude and stabilizes the optimization process. The update rule is given by:

\begin{equation}
\boldsymbol{\theta}_{t+1} = \boldsymbol{\theta}_{t} - \text{lr}_t \ast sign(\textbf{g}_t)
\end{equation}

\noindent where $\boldsymbol{\theta}$ denotes the learnable parameters, including $\alpha$, $\beta$, and $\mathbf{V}$.

Nevertheless, AutoRound is mainly designed for integer-based quantization formats, such as INT4 and INT8. Its learnable scale and rounding adjustments are defined around affine integer quantization. This formulation is not applicable to HiF4 due to its BFP format. In HiF4, the quantized values are determined jointly by the S1P2 element and multiple levels of shared scaling metadata, making the rounding and scaling processes fundamentally different from integer-based quantization. Therefore, we propose HiAutoRound, a calibration method that exploits the characteristics of BFP, enabling learnable parameters to be incorporated into the HiF4 quantization procedure.

\begin{figure*}[t]
    \centering
    \includegraphics[width=\linewidth]{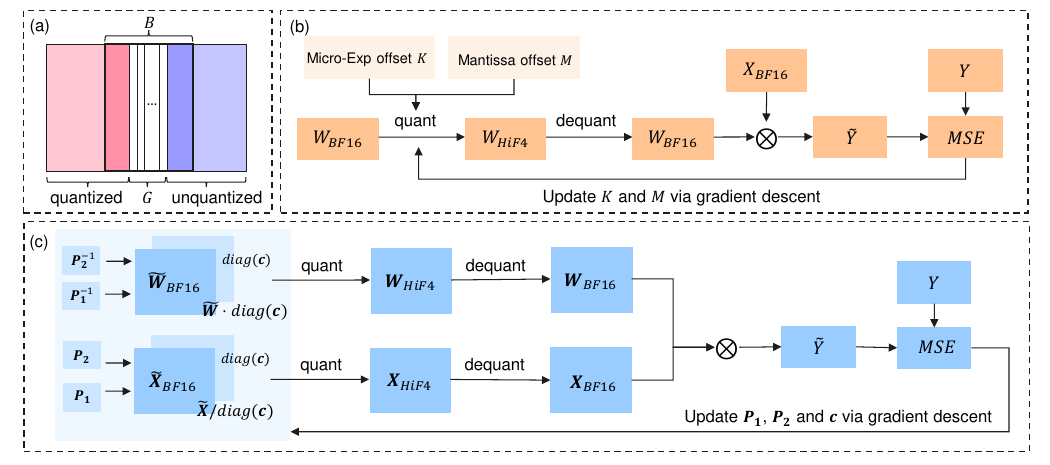}
    \caption{Illustration of the proposed calibration algorithms. (a) \textbf{HiGPTQ}: columns in white represent the group currently being quantized; (b) \textbf{HiAutoRound}: learnable parameters are introduced to automatically adjust the micro-exponent and mantissa components, which are optimized via gradient descent; (c) \textbf{HiFlatQuant}: learnable transformation matrices $\mathbf{P_1}$, $\mathbf{P_2}$ and per-channel scaling $\boldsymbol{c}$ are introduced to flatten outliers in weights and activations, which are optimized via gradient descent.}
    \label{fig:method}
\end{figure*}

\subsubsection{FlatQuant}

FlatQuant is motivated by the observation that quantization errors are often amplified by outliers in weights and activations \cite{zhu2024compression_survey}. 
To mitigate this issue, FlatQuant learns affine transformations for each linear layer to flatten the distributions of weights and activations before quantization. Mathematically, it solves the following problem:

\begin{equation}
\mathbf{P}^{*} = \arg\min_{\mathbf{P}} \left\| \mathbf{Y} - Q(\mathbf{X}\mathbf{P})\, Q(\mathbf{P}^{-1}\mathbf{W}^{\top}) \right\|_{F}^{2}\textbf{}
\end{equation}

\noindent where $\mathbf{Y}$ denotes the original full-precision output, $Q(\cdot)$ denotes the quantization function, and $\mathbf{P} \in \mathbb{R}^{n \times n}$ is the learnable affine transformation matrix. To reduce the computational cost of learning a large dense matrix, FlatQuant decomposes $\mathbf{P}$ into two lightweight matrices using the Kronecker product, i.e., $\mathbf{P}=\mathbf{P_1} \otimes \mathbf{P_2}$, where $\mathbf{P_1}\in \mathbb{R}^{n_1 \times n_1}$, $\mathbf{P_2}\in \mathbb{R}^{n_2 \times n_2}$ and $n=n_1 n_2$. This decomposition significantly reduces the number of learnable parameters while maintaining effective to reshape the weight and activation distributions.

Additionally, FlatQuant employs per-channel scaling to balance outliers between weights and activations, and introduces learnable weight clipping (LWC) and learnable activation clipping (LAC) parameters, denoted by $\alpha_w$ and $\alpha_a$, to further suppress extreme values.
Ultimately, FlatQuant solves the following reconstruction problem:

\begin{equation}
    \min_{\mathbf{P}, \boldsymbol{c}, \alpha_w, \alpha_c} \left\| F_l(\mathbf{X}) - \hat{F}_l(\mathbf{X};\mathbf{P}, \boldsymbol{c}, \alpha_w, \alpha_c) \right\|_{F}^{2}
\end{equation}

\noindent where $F_l{(\cdot)}$ and $\hat{F}_l{(\cdot)}$ denote the original and the quantized Transformer block. 

The learnable clipping parameters $\alpha_w$ and $\alpha_a$ play an important role in FlatQuant, as integer-based low-bit formats such as INT4 have a limited linear representational range and cannot effectively accommodate severe outliers. Adaptive clipping is therefore required to remove or suppress extreme values before quantization. 
In contrast, HiF4 has a substantially larger non-linear dynamic range, which makes it more naturally aligned with the distribution of LLM weights. As a result, HiF4-based calibration does not require learning the extra $\alpha_w$ and $\alpha_a$ parameters, which reduces calibration complexity.

\section{HiF4 Quantization Calibration}

In this section, we formally introduce the proposed calibration methods, namely HiGPTQ, HiAutoRound, and HiFlatQuant, which exploit the structural characteristics of HiF4, enabling more compatible and accurate LLM quantization into the HiF4 format.

\subsection{HiGPTQ}

\begin{algorithm}[!t]
\caption{HiGPTQ}
\label{alg:HiGPTQ}

\begin{algorithmic}[1]

\REQUIRE Weight matrix $\mathbf{W}$, inverse Hessian
$\mathbf{H}^{-1} = (2\mathbf{X}\mathbf{X}^{\top} + \lambda \mathbf{I})^{-1}$,
block size $B$, group size $G=64$ with $B \bmod G = 0$

\ENSURE Quantized weight matrix $\mathbf{Q}$

\STATE $\mathbf{Q} \leftarrow \mathbf{0}_{d_{\mathrm{row}} \times d_{\mathrm{col}}}$
\STATE $\mathbf{E} \leftarrow \mathbf{0}_{d_{\mathrm{row}} \times B}$

\STATE $\mathbf{H}^{-1} \leftarrow \mathrm{Cholesky}(\mathbf{H}^{-1})^{\top}$

\FOR{$i = 0, B, 2B, \dots$}

    \FOR{$g = i, i+G, i+2G, \dots, i+B-G$}

        \STATE $\mathcal{G} \leftarrow g:(g+G)$
        \STATE $\widetilde{\mathcal{G}} \leftarrow (g-i):(g-i+G)$

        \STATE
        $\mathbf{Q}_{:,\mathcal{G}} \leftarrow
        \mathrm{quant_{HiF4}}(\mathbf{W}_{:,\mathcal{G}})$

        \STATE
        $\mathbf{E}_{:,\widetilde{\mathcal{G}}}
        \leftarrow
        \left(
        \mathbf{W}_{:,\mathcal{G}}
        -
        \mathbf{Q}_{:,\mathcal{G}}
        \right) / \mathbf{H}_{\mathcal{G},\mathcal{G}}^{-1}$

        \STATE
        $\mathbf{W}_{:,g:(i+B)}
        \leftarrow
        \mathbf{W}_{:,g:(i+B)}
        -
        \mathbf{E}_{:,\widetilde{\mathcal{G}}} \cdot
        \mathbf{H}_{\mathcal{G},\,g:(i+B)}^{-1}$

    \ENDFOR

    \STATE
    $\mathbf{W}_{:,(i+B):}
    \leftarrow
    \mathbf{W}_{:,(i+B):}
    -
    \mathbf{E} \cdot \mathbf{H}_{i:(i+B),\,(i+B):}^{-1}$

\ENDFOR

\end{algorithmic}
\end{algorithm}

As HiF4 employs a group size of 64, with the level-1 scaling metadata shared across all 64 elements, the original GPTQ's column-wise quantization is not directly suitable. Therefore, HiGPTQ requires the block size to be a multiple of 64, and weights are quantized group by group, as shown in Fig.~\ref{fig:method} (a). In each group, 64 columns are processed simultaneously so that the shared HiF4 scaling metadata can be derived based on all elements in the group. This design better aligns the quantization procedure with the hierarchical structure of HiF4 and allows the group-level scaling factors to be derived more effectively.

Algorithm \ref{alg:HiGPTQ} illustrates the whole procedure of HiGPTQ. Lines 1 and 2 initialize the quantized weight and error matrices. Line 3 pre-computes the Cholesky decomposition factors of the inverse Hessian. Line 4 iterates over each block of the weight matrix, while Line 5 iterates over each HiF4 group within the current block. For each group, HiGPTQ first quantizes the corresponding weights into the HiF4 format and then computes the quantization error between the original weights and the quantized weights. This error is subsequently propagated to the remaining unquantized weights within the same block using the pre-computed Hessian inverse information. After all groups within a block have been processed, Line 12 performs a global update to the remaining weight matrix using the lazy batch-update strategy.

It is worth-noting that, due to the three-level scaling hierarchy of HiF4, the group size can also be set as 8 or 4, corresponding to the level-2 and level-3 scaling granularities, respectively. In these cases, after each group of 8 or 4 columns is quantized, the remaining weights in the block are updated using the same Hessian-based error compensation mechanism. A smaller group size enables more frequent error updates and may provide finer-grained correction. In this paper, we employ the first strategy where 64 columns are processed simultaneously.

\subsection{HiAutoRound}

Inspired by AutoRound which introduces learnable parameters to adaptively adjust rounding operations for integer-based quantization, we propose a learnable calibration mechanism tailored to the floating-point structure of HiF4. Each HiF4 unit comprises 16 shared level-3 micro-exponents and 64 individual S1P2 element. Therefore, we introduce two sets of learnable parameters, denoted as $\mathbf{K}$ and $\mathbf{M}$, to compensate the exponent and mantissa components, respectively.

For exponent compensation, $\mathbf{K}$ is added to the level-3 micro-exponents. Since each level-3 micro-exponent is shared by 4 elements, a HiF4 block of 64 elements contains 16 such micro-exponents. Therefore, we define $\mathbf{K} \in \mathbb{R}^{16}$ with each element corresponding to one level-3 micro-exponent. The compensated exponent is computed as:

\begin{equation}
    E'=E+\operatorname{RZ}(K)
\end{equation}

\noindent where $\operatorname{RZ}$ denotes the rounding-to-zero operation, ensuring that the adjusted exponent remains compatible with the discrete power-of-two scaling structure of HiF4. 

Similarly, the mantissa compensation $\mathbf{M}$ is added to the individual S1P2 element, where $\mathbf{M} \in \mathbb{R}^{64}$. Since $\mathbf{K}$ determines the effective exponent offset, it also changes the numerical scale at which the mantissa compensation is interpreted. Therefore, directly applying $\mathbf{M}$ may introduce discontinuous changes in the reconstructed value when the exponent compensation change. To address this issue, we propose to apply a right-shift operation to $\mathbf{M}$ according to the value of $\mathbf{K}$, as shown in Eq.~\eqref{eq:mantissa_compensation}, to make the mantissa compensation aligned with the adjusted exponent level. During calibration, numerical constraints may be applied to $\mathbf{K}$ and $\mathbf{M}$ to ensure that the compensated values remain consistent with the data distributions of the exponent and mantissa components in the HiF4 format.

\begin{equation}
M' = M \gg \operatorname{RZ}(K)
\label{eq:mantissa_compensation}
\end{equation}

Consequently, each reconstructed value can be expressed as:

\begin{equation}
\begin{aligned}
V_i =
&\ \mathrm{E6M2}
\times
2^{\left(
\{E1\_8\}_{\lceil i/8 \rceil}
+
\{E1\_16\}_{\lceil i/4 \rceil}
+
\{K\}_{\lceil i/4 \rceil} 
\right) }
\times \\
& (\{S1P2\}_i + \{M\}_i \gg \operatorname{RZ}(\{K\}_{\lceil i/4 \rceil}))
\end{aligned}
\label{eq:HiAutoRound_V}
\end{equation}

Algorithm \ref{alg:HiAutoRound} illustrates the full process of HiAutoRound. For each layer to be quantized, the input activations $\mathbf{X}$ are first collected through forward propagation using calibration data. HiAutoRound then quantizes the weights into the HiF4 format according to Eq.~\eqref{eq:HiAutoRound_V}. During calibration, the learnable parameters $\mathbf{K}$ and $\mathbf{M}$ are optimized by minimizing the reconstruction error between the full-precision layer output and the HiF4-quantized layer output using the SignSGD optimizer, as shown in Fig.~\ref{fig:method} (b).

\begin{algorithm}[t]
\caption{HiAutoRound}
\label{alg:HiAutoRound}
\begin{algorithmic}[1]

\REQUIRE Weight matrix $\mathbf{W}$, calibration data $\mathcal{D}$, learning rate $lr$, total steps $T$, batch size $bs$.

\ENSURE $\mathbf{K}^*$, $\mathbf{M}^*$

\STATE $\mathbf{K}\gets 0$, $\mathbf{M}\gets 0$, $\text{min\_}\ell\gets\text{maximum}$
\FOR{$i\gets 0$ to $T$}
    \STATE $d \gets \text{draw} \ bs \  \text{samples from} \ \mathcal{D}$
    \STATE $\mathbf{X} \gets \operatorname{feedforward}({d})$
    \STATE $\widetilde{\mathbf{W}} \gets \operatorname{quant}_{HiF4}(\mathbf{W}, \mathbf{K}, \mathbf{M})$ 
    \STATE $\text{loss}\gets \operatorname{mse}(\mathbf{X}\widetilde{\mathbf{W}}^\top, \mathbf{X}\mathbf{W}^\top)$ 
    \STATE $\text{loss.backward}()$
    \IF{$\text{loss} < \text{min\_}\ell$}
        \STATE $\mathbf{K}^* , \mathbf{M}^*,\gets \mathbf{K}, \mathbf{M}$
        \STATE $\text{min\_}\ell\gets\text{loss}$
    \ENDIF
    \STATE update $\mathbf{K}$ and $\mathbf{M}$ via the SignSGD optimizer 
\ENDFOR
\end{algorithmic}
\end{algorithm}

\begin{algorithm}[t]
\caption{HiFlatQuant}
\label{alg:HiFlatQuant}
\begin{algorithmic}[1]

\REQUIRE Weight matrix $\mathbf{W}$, calibration data $\mathcal{D}$, learning rate $lr$, total steps $T$, batch size $bs$.

\ENSURE $\mathbf{P}_1^*$, $\mathbf{P}_2^*$ and $\boldsymbol{c}^*$ 

\STATE Initialize $\mathbf{P}_1$, $\mathbf{P}_2$, $\boldsymbol{c}$ and $\text{min}\_l$
\FOR{$i\gets 0$ to $T$}
    \STATE $d \gets \text{draw} \ bs \  \text{samples from} \ \mathcal{D}$
    \STATE $\mathbf{X} \gets \operatorname{feedforward}({d})$
    \STATE $\mathbf{Y} \gets \mathbf{X}\mathbf{W}^\top$
    \STATE $\mathbf{X}' \gets \mathbf{P}_{1}^\top\times_{1}\widetilde{\mathbf{X}}\times_{2}\mathbf{P}_{2}$
    \STATE $\mathbf{W}' \gets \mathbf{P}_{1}^{-1}\times_{1}\widetilde{\mathbf{W}}\times_{2}(\mathbf{P}_{2}^{-1})^{\top}$
    \STATE $\widetilde{\mathbf{Y}} \gets \operatorname{quant}_{HiF4}(\mathbf{X}') \times \operatorname{quant}_{HiF4}(\mathbf{W}')^\top$ 
    \STATE $loss \gets \operatorname{mse}(\widetilde{\mathbf{Y}}, \mathbf{Y})$
    \STATE $\text{loss.backward}()$
    \IF{$\text{loss} < \text{min\_}\ell$}
        \STATE $\mathbf{P}_1^*, \mathbf{P}_2^*, \boldsymbol{c}^* \gets \mathbf{P}_1, \mathbf{P}_2, \boldsymbol{c}$
        \STATE $\text{min\_}\ell\gets\text{loss}$
    \ENDIF
    \STATE update $\mathbf{P}_1$, $\mathbf{P}_2$ and $\boldsymbol{c}$ via the AdamW optimizer 
\ENDFOR
\end{algorithmic}
\end{algorithm}

\begin{table*}[th!]
\centering
\caption{Experimental results on diverse models and tasks.}
\label{tab:experiment}
\begin{tabular}{
  >{\centering\arraybackslash}p{1.7cm}   
  >{\centering\arraybackslash}p{2.4cm}   
  >{\centering\arraybackslash}p{1.1cm}   
  >{\centering\arraybackslash}p{1.1cm}   
  >{\centering\arraybackslash}p{1.0cm}   
  >{\centering\arraybackslash}p{1.0cm}   
  >{\centering\arraybackslash}p{1.0cm}   
  >{\centering\arraybackslash}p{1.0cm}   
  >{\centering\arraybackslash}p{1.0cm}   
  >{\centering\arraybackslash}p{2.0cm}   
}
\hline
\multirow{2}{*}{\textbf{Model}} 
& \multirow{2}{*}{\textbf{Quantization}} 
& \multicolumn{7}{c}{\textbf{Datasets}} 
& \multirow{2}{*}{\textbf{Average}} \\
\cline{3-9}
& & ARC-C & ARC-E & BoolQ & HellaS & LamOpai & Piqa & WinoG & \\
\hline

\multirow{7}{*}{Llama2-7B} 
& BF16 & 46.25 & 74.62 & 77.92 & 75.99 & 73.63 & 78.84 & 68.82 & 70.87 \\
& MXFP4 & 40.36 &	66.12 &	72.14 &	71.15 &	69.34 &	77.31 	& 63.54 &	65.71 (-5.16)
 \\
& NVFP4 & 42.75 &	72.90 &	75.35 &	74.59 &	73.47 &	78.07 &	68.75 &	69.41 (-1.46)
 \\
& HiF4 & 44.03 &	73.44 &	76.76 &	74.94 &	72.91 &	78.35 	& 67.56 &	69.71 (-1.16)
 \\
& HiGPTQ & 44.88 &	73.74 &	76.97 &	75.31 &	73.18 &	78.35 &	68.59 &	\textbf{70.15 (-0.72)} \\
& HiAutoRound & 45.31 &	73.15 &	77.06 &	75.19 &	73.24 &	78.56 &	68.35  & \textbf{70.12 (-0.75)} \\
& HiFlatQuant & 45.48 &	73.70 &	77.71 &	74.59 &	72.37 &	78.40 &	68.35 & \textbf{70.09 (-0.78)} \\

\hline

\multirow{7}{*}{Llama2-13B} 
& BF16 & 49.06 &	77.48 &	80.46 &	79.38 &	76.71 &	80.79 &	72.06 &	73.71  \\
& MXFP4 & 44.45 &	72.64 &	77.80 &	73.43 &	73.32 &	77.20 &	69.38 &	69.75 (-3.96)
 \\
& NVFP4 & 47.44 &	75.93 &	79.54 &	77.30 &	75.22 &	79.87 &	71.19 &	72.36 (-1.35)
 \\
& HiF4 & 48.21 &	76.22 &	79.76 &	77.89 &	75.43 &	80.25 	& 71.27 &	72.72 (-0.99)
 \\
& HiGPTQ & 49.49 &	77.36 &	81.44 &	79.15 &	76.65 &	80.14 &	72.77 & \textbf{73.86(+0.15)} \\
& HiAutoRound & 50.94 &	77.27 &	79.79 &	78.85 &	75.88 &	79.98 &	71.19 & \textbf{73.41 (-0.30)} \\
& HiFlatQuant & 50.94 &	77.69 &	80.03 &	78.23 &	76.25 &	80.58 &	72.69 & \textbf{73.77 (+0.06)} \\

\hline

\multirow{7}{*}{Llama3-8B} 
& BF16 & 53.75 & 77.90 &	81.07 &	79.18 &	75.72 &	80.69 &	73.24 &	74.51 \\
& MXFP4 & 45.22 &	72.47 &	72.81 &	73.84 &	68.17 &	78.24 	& 67.56 &	68.33 (-6.18) \\
& NVFP4 & 47.53 &	72.98 &	74.50 &	76.30 &	73.63 & 78.73 &	70.48 &	70.59 (-3.92) \\
& HiF4 & 50.60 &	76.26 &	80.40 &	77.86 &	73.47 &	78.94 	& 73.72 &	73.04 (-1.47) \\
& HiGPTQ & 52.99 &	76.56 &	80.80 &	78.76 &	75.61 &	81.07 &	72.53 & \textbf{74.05 (-0.46)} \\
& HiAutoRound & 50.68 &	76.35 &	80.76 &	78.21 &	75.63 &	80.36 &	74.35 & \textbf{73.76 (-0.75)} \\
& HiFlatQuant & 53.16 &	77.53 &	80.49 &	77.74 &	73.53 &	79.49 &	73.88 & \textbf{73.69 (-0.82)} \\



\hline

\multirow{7}{*}{Qwen2.5-7B} 
& BF16 & 51.28 & 77.69 & 84.62 &	78.92 &	71.74 &	79.65 &	73.01 &	73.84  \\
& MXFP4 & 50.17 & 75.93 & 79.66 &	73.75 &	58.84 &	78.56 &	66.22 &	69.02 (-4.82) \\
& NVFP4 & 50.43 & 77.78 & 84.43 &	77.36 &	68.17 &	79.33 	& 69.46 &	72.42 (-1.42) \\
& HiF4 & 49.66 & 76.60 & 82.51 &	77.57 &	66.76 &	79.33 	& 71.43 &	71.98 (-1.86) \\
& HiGPTQ & 51.28 & 76.35 &	83.39 &	78.07 &	71.07 &	80.30 &	73.32 & \textbf{73.40 (-0.44)} \\
& HiAutoRound & 51.45 &	76.94 &	83.06 &	77.70 &	71.34 &	79.22 &	72.61 & \textbf{73.19 (-0.65)} \\
& HiFlatQuant & 52.65 &	78.24 &	84.22 &	77.06 &	68.52 &	78.13 &	72.06 & \textbf{72.98 (-0.86)} \\

\hline

\multirow{7}{*}{Qwen2.5-14B} 
& BF16 & 59.13 & 79.34 &	85.41 &	82.93 &	74.17 &	82.32 &	75.14 &	76.92 \\
& MXFP4 & 55.72 &	79.71 &	84.53 &	79.37 &	68.93 &	81.01 	& 71.90 &	74.45 (-2.47) \\
& NVFP4 & 57.85 &	81.31 &	84.10 &	81.31 &	72.79 &	81.45 	& 75.45 &	76.32 (-0.60) \\
& HiF4 & 58.70 &	82.03 &	85.93 &	81.48 &	72.71 &	81.61 &	73.56 &	76.57 (-0.35) \\
& HiGPTQ & 60.41 &	80.77 &	85.60 &	82.42 &	74.66 &	81.77 &	73.95 & \textbf{77.08 (+0.16)} \\
& HiAutoRound & 57.94 &	81.23 &	85.87 &	81.43 &	72.54 &	81.99 &	75.37 & \textbf{76.62 (-0.30)}	\\
& HiFlatQuant & 59.90 &	81.82 &	84.74 &	82.08 &	73.53 & 81.61 &	75.85 &	\textbf{77.08 (+0.16)} 
 \\

\hline

\end{tabular}
\end{table*}

\subsection{HiFlatQuant}

As shown in Fig.~\ref{fig:method} (c), HiFlatQuant aims to flatten outliers in weights and activations by introducing a learnable transformation matrix $\mathbf{P} \in \mathbb{R}^{n \times n}$ and further decomposing $\mathbf{P}$ into the Kronecker Product of two lightweight matrices, $\mathbf{P}_1 \in \mathbb{R}^{n_1 \times n_1}$ and $\mathbf{P}_2 \in \mathbb{R}^{n_2 \times n_2}$, where $n=n_1n_2$. Accordingly, the transformed quantized computation can be written as \cite{sun2024FlatQuant}:

\begin{equation}
\begin{aligned}
& Q(\mathbf{X} \mathbf{W}^\top) =
Q(\mathbf{X}\mathbf{P})Q(\mathbf{P}^{-1}\mathbf{W}^{\top}) \\
& =
\ Q\left( \mathbf{P}_{1}^\top\times_{1}\widetilde{\mathbf{X}}\times_{2}\mathbf{P}_{2}\right)
\times
Q\left(\mathbf{P}_{1}^{-1}\times_{1}\widetilde{\mathbf{W}}\times_{2}(\mathbf{P}_{2}^{-1})^{\top}\right)^{\top},
\end{aligned}
\end{equation}

\noindent where $\widetilde{\mathbf{X}} \in \mathbb{R}^{k \times n_1 \times n_2}$ and $\widetilde{\mathbf{W}} \in \mathbb{R}^{m \times n_1 \times n_2}$ are reshaped from $\mathbf{X}$ and $\mathbf{W}$ accordingly, and $\times_i$ denotes the reduction over the $i$-th axis.

\begin{table*}[t]
\centering
\caption{Average performances across models and tasks.}
\label{tab:accuracy_average}
\begin{tabular}{lccccccc}
\hline
\textbf{Metric} 
& \textbf{BF16} 
& \textbf{MXFP4} 
& \textbf{NVFP4} 
& \textbf{HiF4} 
& \textbf{HiGPTQ} 
& \textbf{HiAutoRound} 
& \textbf{HiFlatQuant} \\
\hline
Accuracy 
& 73.970 & 69.452 & 72.220 & 72.804 & 73.708 & 73.420 & 73.522 \\
Accuracy gap 
& -- & -4.518 & -1.750 & -1.166 & -0.262 & -0.550 & -0.448 \\
\hline
\end{tabular}
\end{table*}

Similar to FlatQuant, HiFlatQuant also applies per-channel scaling $\boldsymbol{c}$ to balance weights and activations before the quantization. However, FlatQuant further introduces learnable weight clipping $\alpha_w$ and activation clipping $\alpha_a$ to suppress remaining outliers. In HiFlatQuant, these additional clipping parameters are not included. This design is motivated by the fact that HiF4 has a substantially larger dynamic range than integer-based low-bit formats. As a result, HiF4 can naturally accommodate a wider range of values, making explicit learnable clipping less necessary. Including $\alpha_w$ and $\alpha_a$ would bring little performance improvement while increasing calibration complexity. Therefore, HiFlatQuant retains the affine transformation and per-channel scaling mechanisms, while omitting learnable clipping to achieve a better trade-off between accuracy and calibration efficiency. The full procedure of HiFlatQuant is illustrated in Algorithm \ref{alg:HiFlatQuant}.

\section{Experiment}

In this section, we evaluate the effectiveness of the proposed HiF4-based calibration methods by addressing the following research questions:

\begin{itemize}
    \item \textbf{RQ1}: Do HiGPTQ, HiAutoRound and HiFlatQuant improve model quantization compared with direct HiF4 casting?
    \item \textbf{RQ2}: How do HiF4-calibrated models perform compared with other state-of-the-art low-precision formats?
    \item \textbf{RQ3}: What are the differences and relationships among the three proposed calibration approaches?
\end{itemize}

\begin{table*}[t]
\centering
\caption{Comparison of three calibration approaches}
\label{tab:comparison}
\begin{tabular}{
    p{2.8cm}
    p{4.2cm}
    p{4.2cm}
    p{4.2cm}
}
\hline
\textbf{Dimension} 
& \textbf{HiGPTQ} 
& \textbf{HiAutoRound} 
& \textbf{HiFlatQuant} \\
\hline

Mechanism
& Compensate quantization error based on Hessian inverse
& Learn offset parameters in micro-exponent and mantissa components
& Learn equivalent transformations to flatten outliers \\

Optimization
& Second-order approximation 
& Gradient-based parameter tuning
& Gradient-based transformation learning \\

Quantization target 
& Weight
& Weight
& Weight and activation \\

Inference overhead 
& None 
& None 
& Depends on transformation folding \\

\hline
\end{tabular}
\end{table*}

\subsection{Experimental settings}

\subsubsection{Datasets}
To evaluate quantization accuracy, we use a series of widely adopted LLMs benchmarks, including ARC-C(hallenge), ARC-E(asy) \cite{clark2018ARC}, BoolQ \cite{clark2019boolq}, HellaS(wag) \cite{zellers2019hellaswag}, Lam(bada)Op(en)ai \cite{paperno2016lambada}, Piqa \cite{bisk2020PIQA}, and WinoG(rande) \cite{sakaguchi2021winogrande}. 

\subsubsection{Models}
We evaluate the proposed calibration methods using various LLMs, including LLaMA2-7B \cite{touvron2023llama2}, LLaMA2-13B \cite{touvron2023llama2}, LLaMA3-8B \cite{grattafiori2024llama3}, Qwen2.5-7B \cite{qwen2025qwen25}, and Qwen2.5-14B \cite{qwen2025qwen25}, which allows us to evaluate the robustness and generalizability of the proposed calibration methods across models.

\subsubsection{Benchmarks}

To demonstrate the effectiveness of the proposed calibration methods, we compare them with direct HiF4 conversion \cite{luo2026hifloat4}, as well as MXFP4 \cite{rouhani2023_MXFP4} and NVFP4 \cite{abecassis2025pretraining_NVFP4} 4-bit FP structures. 

\subsubsection{Implementation}

For calibration data, we sample 128 sentences from WikiText2 \cite{merity2017wikitext2} training set, with each sample comprised of 2048 tokens. For HiGPTQ, we set the group size to $G=64$ and block size to $B=64$. For HiAutoRound, we employ the SignSGD optimizer with a learning rate of 0.001 and set the number of iteration to be 100. For HiFlatQuant, we adopt the AdamW optimizer with an initial learning rate of 1e-3 and employ a learning rate decay schedule. The number of iterations is set to 5. Batch sizes are adjusted according to model sizes. All experiments are conducted in the simulated 4-bit format on Huawei Ascend NPU. 
To ensure a fair comparison, all evaluations are performed in a unified framework with the same prompt for model inference.

\subsection{Results (RQ1 and RQ2)}

The detailed evaluation results are presented in Table \ref{tab:experiment}, and the average accuracies across all models and tasks are summarized in Table \ref{tab:accuracy_average}. Notably, direct HiF4 casting already achieves higher accuracy than both MXFP4 and NVFP4, demonstrating the inherent representational advantage of HiF4. This advantage can be attributed to its three-level scaling hierarchy, which provides a substantially broader dynamic range. 

Across all models and datasets, HiGPTQ, HiAutoRound and HiFlatQuant achieve average accuracies of 73.708\%, 73.420\% and 73.522\%, respectively, corresponding to improvements of 0.904, 0.616 and 0.718 percentage points over direct HiF4 casting.
This reduces the accuracy gap between HiF4 and BF16 to less than 1\%, highlighting the effectiveness of the proposed calibration methods in preserving model accuracy under 4-bit quantization.

\subsection{Comparative analysis (RQ3)}

Evaluation results demonstrate the effectiveness of all three proposed calibration methods for HiF4 quantization. Nevertheless, these methods differ in their underlying mechanisms, optimization strategies, quantization targets and inference overload. Table \ref{tab:comparison} summarizes these differences, providing a more comprehensive understanding of their characteristics and suitable usage scenarios.

\section{Conclusion}
In this paper, we propose three HiF4-based calibration algorithms, namely HiGPTQ, HiAutoRound and HiFlatQuant, for large model quantization. The proposed methods adapt existing calibration principles to the hierarchical BFP structure of HiF4 through distinct mechanisms. HiGPTQ leverages second-order information to compensate for quantization errors by updating the remaining unquantized weights. HiAutoRound introduces learnable offset parameters to adaptively adjust the micro-exponent and mantissa components. HiFlatQuant learns equivalent transformation matrices to flatten outliers in weights and activations before quantization. Experimental results demonstrate that all three calibration methods improve quantization accuracy over direct HiF4 casting. Moreover, the calibrated HiF4 models reduce the performance gap from the high-precision BF16 baseline to less than 1\%, outperforming the state-of-the-art NVFP4 format across 5 models and 7 downstream tasks.

\bibliography{aaai2027}


\end{document}